\documentclass[screen,nonacm,11pt]{acmart}

\makeatletter
\providecommand*{\input@path}{}
\edef\input@path{{styles/}{../styles/}\input@path}
\makeatother

\RequirePackage{ragged2e}
\usepackage{AuxCommands}
\usepackage{subfiles}

\DeclareGraphicsExtensions{.pdf,.png,.jpg}
\graphicspath{{images/pdf/}{../images/pdf/}{images/tikz/}{../images/tikz/}{images/png/}{../images/png/}}

\begin{document}
\title[CNN Compression]{Data-Driven Compression of Convolutional Neural Networks}

\author{Ramit~Pahwa}
\email{rapahwa@adobe.com}
\affiliation{%
	\institution{Adobe}
}
\author{Manoj~Ghuhan~Arivazhagan}
\email{ghuhana@adobe.com}
\affiliation{%
	\institution{Adobe Research}
}
\author{Ankur~Garg}
\email{ankgarg@cs.utexas.edu}
\affiliation{%
	\institution{The University of Texas at Austin, USA}
}
\author{Siddarth~Krishnamoorthy}
\email{siddarrth.k@gmail.com}
\affiliation{%
	\institution{Indian Institute of Technology, Kanpur, India}
}
\author{Rohit~Saxena}
\email{saxenarrohit@gmail.com}
\affiliation{%
	\institution{Indian Institute of Technology, Roorkee, India}
}
\author{Sunav~Choudhary}
\email{schoudha@adobe.com}
\affiliation{%
	\institution{Adobe Research}
}

\renewcommand{\shortauthors}{Pahwa et al.}

\begin{abstract}
    Deploying trained convolutional neural networks (CNNs) to mobile devices is a challenging task because of the simultaneous requirements of the deployed model to be fast, lightweight and accurate.
    Designing and training a CNN architecture that does well on all three metrics is highly non-trivial and can be very time-consuming if done by hand.
    One way to solve this problem is to compress the trained CNN models before deploying to mobile devices.
    This work asks and answers three questions on compressing CNN models automatically:
    a) How to control the trade-off between speed, memory and accuracy during model compression?
    b) In practice, a deployed model may not see all classes and/or may not need to produce all class labels.
    Can this fact be used to improve the trade-off?
    c) How to scale the compression algorithm to execute within a reasonable amount of time for many deployments?
    The paper demonstrates that a model compression algorithm utilizing reinforcement learning with architecture search and knowledge distillation can answer these questions in the affirmative.
    Experimental results are provided for current state-of-the-art CNN model families for image feature extraction like VGG and ResNet with CIFAR datasets.
\end{abstract}



\maketitle
\renewcommand{\bibsubfile}[2]{}

\section{Introduction}
    \label{sec:intro}
    Convolutional Neural Networks (CNNs) have enabled monumental progress in many computer vision tasks over the past five years, achieving and even surpassing human level cognition~\cite{szegedy2017inception,real2018regularized,hu2017squeeze}.
    Such networks are often carefully designed and have become very deep and large, especially the ones that claim to achieve state-of-the-art results.
    For example, the ImageNet Large Scale Visual Recognition Challenge (ILSVRC) for the year 2015 was won by a deep neural network from the ResNet~\cite{he2015deep} family with 152 layers and obtained a 3.57\% error rate.
    Unfortunately, very deep networks can have a large memory footprint and become slow during inference.
    While this may not be a big problem when the trained model is deployed as a cloud service, it becomes a deal breaker for deployments on to the vast majority of comparatively resource constrained mobile devices and embedded systems.
    
    Compressing trained CNNs has been suggested as one avenue to tame very large and deep networks for the purpose of deployment to resource constrained devices.
    While there has been significant progress in this direction recently (see the survey in~\cite{cheng2017survey}), it remains an active area of research with several open questions and challenges.
    One observation made in~\cite{cheng2017survey} is that the various approaches to compressing CNNs are somewhat orthogonal which raises the question ``What are the right principles to employ for compressing families of CNN model architectures?''
    Yet another question is whether smaller/shallower neural networks with faster inference can come (robustly) close to the state-of-the-art results achieved by the deep CNN families, and if this could be done without the trial and error overhead of hand designed model architectures.
    In this work, we pose and study three related questions:
    \begin{enumerate}
    	\item	For deploying models to resource constrained devices, the accuracy, memory footprint and inference speed of the deployed model are important considerations.
    	How do we control the trade-off between these quantities (henceforth, referred to as the `AMS trade-off')?
    	\item	In most applications, a deployed model will only see a reduced diversity of input data and/or will need to produce only a subset of the possible class labels.
    	Can this aspect be used to improve the AMS trade-off?
    	As a crude example, consider a neural network deployed on a self-driving car for identifying objects in its path.
    	Such a network might have been trained on the ImageNet dataset, but it only needs to be able to classify objects into two categories, viz. objects for which it needs to slow down or stop, and everything else.
    	\item	Current algorithms for model compression are fairly compute intensive.
    	Depending on the particular algorithm and the compression targets that need to be achieved, it could take anywhere from a few hours to a couple of days to get a satisfactory result.
    	If several different compressed models need to be generated, is it possible to reduce the amortized execution time?
    	As a real-world example, consider a deep neural network that needs to power the user experience on a mobile application.
    	Since users have different behavioral patterns, there is a case for deploying compressed models that are personalized to users/user segments.
    	Scaling such a `personalized compression' task beyond a few users becomes prohibitive very quickly.
    \end{enumerate}
    
    Our contributions are to help answer the above questions and we do so by developing an algorithm for compressing CNNs.
    We draw upon prior work in~\cite{ashok2017n2n} to utilize reinforcement learning (RL) to learn a compression policy for a given CNN trained on a given dataset, where compression is via architecture search and the reward function includes compression rate and accuracy of a candidate architecture (student model) trained using knowledge distillation~\cite{hinton2015distilling,romero2014fitnets} from the given CNN (teacher model).
    We name our method as Data-driven Compression (DDC) and add the following over and above the contributions in~\cite{ashok2017n2n}.
    \begin{enumerate}
	    \item	Besides compression rate and accuracy, we include inference time as a component of the reward function for RL.
	    Further, we design reward functions with user definable performance thresholds for each of the three metrics in the AMS trade-off.
	    The thresholds allow for control of the operating point in the AMS trade-off space for different deployment targets.
	    \item	We demonstrate that the compression policy learned on the full dataset generalizes to the compression task \wrt~data subsets having fewer class labels.
	    In other words, using the compression policy learned on the full dataset gives a \emph{better} AMS trade-off on the compression task \wrt~a data subset than executing the compression from scratch with a policy learned on the data subset.
	    Moreover, using the former compression policy leads to much faster completion (5x faster) of the compression task \wrt~data subsets than compressing from scratch.
	    This leads to much smaller amortized execution time as compared to other model compression algorithms when multiple compressed models need to be generated while maintaining an impressive AMS trade-off for each compressed model.
	    We conduct extensive experiments with ResNet and VGGnet model families trained on CIFAR datasets to demonstrate these results.
	    Finally, we note that the compression policy transfer claims and supporting experiments in~\cite{ashok2017n2n} are of an orthogonal nature.
	    Therein, it is claimed that the learned compression policy generalizes to the compression task on other models from the same architecture family, \eg~compression policy learned on ResNet-18 can generalize to compress ResNet-34 models.%
    \end{enumerate}
    


\bibsubfile{named}{bibs/sub}

\section{Related Work}
    \label{sec:related}
    Reducing the depth of the networks \cite{iandola2016squeezenet} and utilizing less expensive operations, such as depth-wise convolutions~\cite{howard2017mobilenets} and group convolutions \cite{cheng2017survey} had gained momentum primarily due to their compactness and ease of deployment to restricted environment.
    These structures are special and hand designed.
    These networks have been almost entirely superseded by Architecture Search \cite{zoph2017learning, jin2018efficient} and Compression \cite{tan2018mnasnet,he2018amc}.
    In this section, we discuss the previous approaches that address the problem of compressing a given network architecture to smaller networks.
    There are mainly three such approaches, \viz~pruning, knowledge distillation, and architecture search.
    
    The pruning approach \cite{lecun1990optimal,srinivas2015data,guo2016dynamic,anwar2017structured} removes the neural network weights that contribute very little towards the performance of the model.
    A known issue with pruning is that it can over compress and damage the network beyond repair \cite{molchanov2016pruning}.
    Further, there are very few human controls in the pruning method.
    In other words, metrics of interest like inference latency, accuracy and compression ratio cannot be directly optimized.
    
    The knowledge distillation approach \cite{hinton2015distilling,romero2014fitnets,ba2014deep} trains a smaller network architecture (student) by utilizing the outputs of the original network (teacher).
    However, this approach is limited by the need to devise the student architecture.

    Given a neural network, the architecture search approach involves searching for a smaller architecture (student) in the teacher architecture space that can display performance close to the original neural network.
    In general, brute force search through smaller architectures is computationally expensive.
    Recently, more principled search methods based on RL have been proposed \cite{zoph2016neural,baker2016designing}.
    Furthermore, design of structured search spaces for good architectures has been undertaken using RL~\cite{zoph2017learning} and using evolutionary algorithms \cite{real2018regularized,real2017large}.
    More recently, Bayesian optimization has been proposed for hyper-parameter tuning~\cite{jin2018efficient}.
    This system (called Auto-Keras) also searches architecture from scratch. Searching an architecture from scratch \cite{zoph2017learning,tan2018mnasnet} has its limitation as it takes unrealistic time to search an optimal architecture for large datasets.
    These methods are limited when considering metrics that need to be controlled when deploying to resource constrained devices.
    Some of these restrictions have been recently incorporated in the architecture search space design \cite{tan2018mnasnet,elsken2018multi,cheng2018searching,ashok2017n2n} to control the trade-off between performance and architectural complexity.
    For example, \cite{cao2019learnable,ashok2017n2n} reduce the search complexity by restricts the student architecture search space to that of the original model's architecture. \cite{cao2019learnable} uses Bayesian optimization to compress the original teacher model to new student architecture instead of searching from scratch.
    We build on this premise restricting our search space to that of the teacher model.
    This is computationally less expensive than searching architectures from scratch.
    We also introduce threshold-ed reward to further enforce search of optimal architecture which resides within the resource restrictions.
    A comparison of the features of different RL based methods for architecture search is provided in \tablename~\ref{tab:methods-comparison}.
    
    \begin{table}[t]
        \centering
        \caption{Feature Comparison of RL based methods for architecture search.
        NAS: Neural Architecture Search~\protect\cite{zoph2017learning}, AMC: AutoML for Model Compression~\protect\cite{he2018amc}, N2N: Network to Network Compression~\protect\cite{ashok2017n2n}, MnasNet: Platform-Aware Neural Architecture Search~\protect\cite{tan2018mnasnet}, DDC: Data Driven Compression}
         \begin{tabular}{lccccc}
            \toprule
            							& NAS			& AMC			& N2N			& MnasNet		& DDC			\\
            \midrule
            Accuracy Optimization		& \checkmark	& \checkmark	& \checkmark	& \checkmark	& \checkmark	\\
            Memory Optimization			&				&				& \checkmark	&				& \checkmark	\\
            Latency Optimization		&				& \checkmark	&				& \checkmark	& \checkmark	\\
            Threshold based Search		&				& 				&				& \checkmark	& \checkmark	\\  
            RNN Policy network			& \checkmark	&				& \checkmark	&				& \checkmark	\\
            Fast exploration on GPU		&				& \checkmark	& \checkmark	& \checkmark	& \checkmark	\\
            Distillation from Teacher	&				&				& \checkmark	&				& \checkmark	\\
            \bottomrule
         \end{tabular}
        \label{tab:methods-comparison}
    \end{table}

\bibsubfile{named}{bibs/sub}

\section{Approach}
 \label{sec:approach}
    Our approach closely follows network to network compression [N2N] work done in \cite{ashok2017n2n}, which introduces compressing down from high performing teacher models by modelling it a Markov Decision Process (MDP).
    Empirically by visualizing the activation's of teacher model, we observe the presence of redundant filter in each layer.
    We aim to automatically find the redundant layers in the network.
    To this extent, we train an RL agent to predict binary actions to keep or remove a layer, then update our agent by encouraging exploration of smaller, faster and more accurate model using thresholded reward functions.
    We systematically reduce the teacher model by sequentially deciding whether to keep an entire layer in the network architecture of the teacher model.
    This process is in contrast to \cite{he2018amc} which utilizes actor-critic network to decide on the fraction of filters to keep in particular.
    Formally, our state space $\setS{S}$ comprises of the all the architectures obtained by compressing the teacher model, \ie~by removing layers from the teacher’s architecture.
    The action set $\setS{A}$ consists of binary decision variable $a_{t} \in \cc{0,1}$, which enables us to control which layer to remove/keep in the teacher network.

    \subsection{Problem Definition}
        \label{sec:num1}
        Compression of deep neural network can be achieved by parameter reduction.
        Parameter reduction can be achieved in two ways: First, is pruning individual unimportant elements in the weight matrix \cite{han2015deep} which achieves high degree of compression while preserving accuracy.
        The only drawback of such algorithms is that it requires specialized hardware such as EIE \cite{han2016eie} as the resultant weight matrix is sparse and irregular.
        On the other hand, structured pruning aims to remove entire regions (\eg~channels, layers, block, \etc).
        The resultant weight matrix is regular and can be accelerated directly with existing hardware and libraries.
        Our goal is reducing our complex teacher architecture, by finding out the irrelevant layer/layers which contribute little to the overall performance of the network.
        Let us consider VGG network \cite{simonyan2014very}, this network is a deep convolutional neural network.
        (VGG network is sequential in nature and typically have varying depth between 11 to 19).
        Now our aim is to remove a convolution layer/layers from the above architecture without perturbing its performance.

    \subsection{Recurrent Policy Network}
        \label{sec:num2}
        We leverage RL to search for an optimal architecture.
        Unlike \cite{zoph2017learning,tan2018mnasnet,zhou2018resource} which searches the optimal architecture from scratch.We restrict our state space to just the Teacher model Architectures (\viz~VGG and Resnet) leading exponential reduction in architecture search time.
        We compress deep neural network by removing redundant layers from the network following which we use heuristic based reduction of filters \cite{molchanov2016pruning} to further decrease the footprint of the model, resulting in a compact student model which has comparable performance to its respective teacher and is more computationally efficient.
        Each layer in the teacher’s architecture is characterized by the following tuple: 
        \begin{equation}
             L_{t} = (t, k, s, p, n, s_{start}, s_{end})
         \end{equation}
        where $t$ represents the current layer under consideration, $k$ is kernel size, $s$ is stride and $n$ is the number of outputs (filters).
        To cater for the presence of residual connections in Resnet~\cite{he2015deep} architecture family, we use two additional parameters $s_{start}$ and $s_{end}$ to explicitly inform the policy network of a skip connection.
        This allows us to distinctly represent the layers of the network under consideration.
        We feed the layer representation $L_{t}$ to a LSTM unit which outputs a hidden network embedding $h_{t}$.
        Here, the action to remove the current layer $L_{t}$ affects the preceding  $L_{t-1}$ and succeeding  $L_{t+1}$ layers thus, we use a Bi-directional LSTM network which is illustrated in \textbf{figure-1}.
        The binary action $a_{t}$ which determines the presence or absence of the current layer depends upon the hidden state from both forward $h_{t}$ and backward direction $h_{t}$ as well as the current layer representation $L_{t}$.
        Formally, the policy is defined as:
        \begin{equation}
                \pi(a_{t}|h_{t-1}, h_{t+1}, x_{t})
        \end{equation}
        where $\pi$ represents the policy.
        The produced network architecture needs to satisfy the user specific device restriction, to model this we propose a \textit{threshold-ed reward function} to specify bounds on the accuracy $a$, size $c$ and inference latency $l$.
        This enables us to search for efficient network architectures in a systematic manner.
        The architectures searched satisfy accuracy requirements while still having realistic size and inference latency.
        The next section describes in detail the construction of the reward function $R$.

    \subsection{Reward Function}
        \label{sec:num3}
        \begin{figure}[t]
			\centering
			\includegraphics[width=0.95\figwidth]{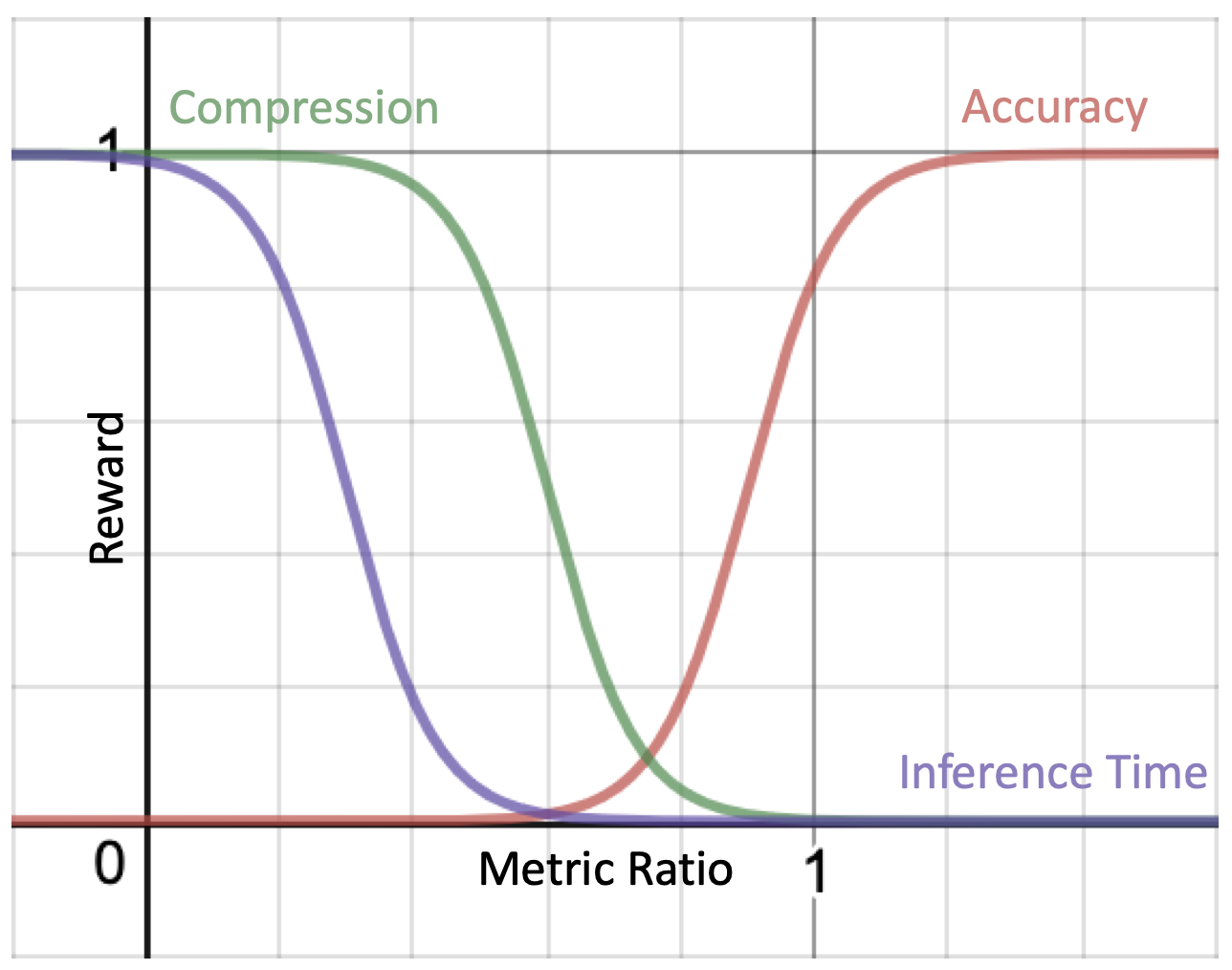}
			\caption{Plot showing the reward structure of the 3 performance metrics}
			\label{fig:rewardfunction}
		\end{figure}
        An intelligently designed reward function which can differentiate between good and bad architectures is necessary for effective exploration in the architectural space.
        We introduce threshold-ed reward function which incorporates user specific device restriction, this allows us to reinforce the policy network to learn architectures which fit the device restrictions.
        Our objective is obtaining architectures which have high accuracy, low memory footprint as well as faster and computationally inexpensive at test time.
        Models having lower inference time (test time) and reasonably high accuracy are preferred over models with very high accuracy (in particular very deep models) which have higher inference time.
        Unlike previous work which optimize for indirect metrics such as FLOPS \cite{he2018amc}, we consider the inference latency on a single GPU as the measure of inference latency.
        We formulate a reward which is a combination three metrics for a given model $m$: Accuracy $A \bb{m}$, Compression percentage $C \bb{m}$ and Inference time $T \bb{m}$ (latency of model $m$) .
        We incorporate user specific device restrictions by introducing thresholds on Accuracy $A \bb{m}$ and Latency $T \bb{m}$.
        Incorporating real time latency in our reward allows us to effective search through the architectural space, resulting in model have high compression and reasonably high accuracy with the added benefit of lower inference time (which is essential element to be considered while deployment of the models to mobile devices)
        We perform intuitive transformation on accuracy Accuracy $A \bb{m}$, Compression percentage $C(m)$ and latency $T\bb{m}$ which are discussed in detail below.\\
        \textbf{Accuracy $A\bb{m}$:} Our primary objective is to obtain architectures which give high performance.
        One way to achieve this is by increasing the depth of the network \cite{he2015deep,simonyan2014very} but this comes at the expense of latency.
        We transform the reward obtained from accuracy such that we obtain a higher reward for model having high accuracy and is continuous function which provides flexibility to incorporate thresholds $A_{th}$ in the reward.
        Formally, reward is obtained through the following transformation.
        $R_1\fcolon A \rightarrow [0,1]$, which normalizes as well as enforces threshold to prevent exploration in undesired region of the search space.
            \begin{equation}
                R_1 \bb{A} = 1 - \inv{\bb{1+\exp\bb{15\cdot\bb{\frac{A}{A_{teacher}}-A_{th}}}}}
            \end{equation}
        where $A_{teacher}$ is accuracy of the teacher model. \\
        \textbf{Latency $T \bb{m}$ :} Inference Latency is essential component to considered which searching for a architecture fit for deployment.
        We incorporate the threshold by using a \textit{shifted sigmoid} transformation which is defined as : \\
        $R_2\fcolon T \rightarrow [0,1]$:
        \begin{equation}
                R_2 \bb{T} = \inv{\bb{1+\exp\bb{15\cdot\bb{\frac{T}{T_{teacher}}-T_{th}}}}}
        \end{equation}
        \textbf{Compression $C \bb{m}$ :} We use the ratio of the number of parameters of student and the teacher model and define the compression index as :\\
        $ C \bb{m} =  \#parameter(student)$, then similarly transform it to enforce our thresholds on size.\\
         \begin{equation}
                R_3 \bb{C} = \inv{\bb{1+\exp\bb{15\cdot\bb{\frac{C}{C_{teacher}}-C_{th}}}}}
        \end{equation}
        Thresholds are needed as we search for the right balance in AMS trade-off Space the thresholds are $A_{th} = 0.9 $ $T_{th} = 0.3 $ $C_{th} = 0.6 $ are fixed for all of our experiments.
        Results pertaining to changing thresholds value are provided in Appendix-\ref{}.
        As \textit{models with high degree of high compression does not guarantee lower inferences time}. The reward for the three performance metrics is given in~\ref{fig:rewardfunction}
        The final reward which need to be maximized for the RL process is defined as:
        \begin{equation}
        R \bb{m} = R_1 \bb{A\bb{m}}\cdot R_2 \bb{T\bb{m}} \cdot R_3 \bb{C\bb{m}}
        \end{equation}

    \subsection{Optimization}
        The parameters of the policy network characterized by $\theta$ are optimized to obtain a efficient policy to compress the teacher model.
        The optimization is formulated to maximize the \textit{expected reward} obtained from the newly compressed architecture defined as :
        \begin{equation}
            \theta^{*} = \arg \max_{\theta} E_{\bb{\Vec{s}, \Vec{a}} \sim p_{\theta\bb{\Vec{s}, \Vec{a}}}} \bb{R\bb{\Vec{s}}}
        \end{equation}
        \begin{equation} 
            J \bb{\theta} = E_{\bb{\Vec{s}, \Vec{a}} \sim p_{\theta\bb{\Vec{s}, \Vec{a}}}} \bb{R\bb{\Vec{s}}}
        \end{equation}
        where $R\bb{\Vec{s}}$ is total reward obtained.
        The optimization process can be estimated using REINFORCE policy gradient used in \cite{williams1992simple}, the continuous nature of our proposed reward transformation improves our search efficiency as it prevents exploding gradient.
        The gradient is estimated as:
        \begin{equation}
            \nabla_{\theta}  J \bb{\theta} = \nabla_{\theta} E_{\bb{\Vec{s}, \Vec{a}} \sim p_{\theta\bb{\Vec{s}, \Vec{a}}}} \bb{R\bb{\Vec{s}}}
        \end{equation}
        \begin{equation}
             \nabla_{\theta}  J \bb{\theta} \approx \frac{1}{N} \sum_{i = 1}^{N} \sum_{t = 1}^{T} \nabla_{\theta} \log{p_{\theta}\bb{a_{i,t}, s_{i,t}}} \bb{R\bb{\Vec{s}}}
        \end{equation}
        here $N$ represents number of produced architectures, $T$ represents the length of trajectory.
        The above equation has high variance, to normalize that we  utilize exponential moving average of the previous rewards as the baseline $b$ subtract it from our total reward $R\bb{\Vec{s}}$.
        \begin{equation}
             \nabla_{\theta}  J \bb{\theta} \approx \frac{1}{N} \sum_{i = 1}^{N} \sum_{t = 1}^{T} \nabla_{\theta} \log{p_{\theta}\bb{a_{i,t}, s_{i,t}}} \bb{R\bb{\Vec{s}} - b}
        \end{equation}
        This helps improve stability of the estimated gradients.

    \subsection{Knowledge Distillation}
    \label{sec:KD}
        Student model architectures are trained utilizing both the outputs of the teacher models and the true label.
        Instead of just using the un-normalized log probabilities (logits) of the teacher model, which outperforms the training process used in \cite{ashok2017n2n}.
        Training incorporating dark knowledge \cite{hinton2015distilling} that helps student to mimic the relationships learned by the teacher model.
        The loss function is trained as combination of \textit{hard} and \textit{soft} targets, giving higher  priority to transferring the dark knowledge.
        If $y_{i}$ are output logits of the teacher model of the $i^{th}$ training example, $y_{true}$ is the true labels.
        Then the loss function is described below as:
        \begin{align}
            \mathcal{L} & = \lambda\cdot\mathcal{L}_{soft} + \bb{1-\lambda}\cdot\mathcal{L}_{hard}   \\
            \mathcal{L}_{soft} & = D_{KL}\bb{f(x; W) \,\Vert\, y_{true}}    \notag \\
            & = \sum_{i} f(x^{i}; W) \cdot \log\frac{f(x^{i}; W)}{y_{true}(i)}   \\
            \mathcal{L}_{hard} & = H\bb{f(x; W), y} = - \sum_{i} f(x^{i}; W) \cdot \log{y_{i}}
        \end{align}
        
        Through experimentation we have fixed value of $\lambda = 0.7$ thus making the student model to mimic the behaviour of the teacher model simultaneously fine-tuning the student architecture towards the true labels.

    \subsection{Transferring Learned Compression Policy}
    \label{sec:TL}
        Unlike in the previous literature \cite{zoph2017learning, zoph2016neural} of transferring knowledge between different architecture families, we show transferability of the learned parameters of the policy across different subsets of data from a policy learned on the entire dataset .
        This not only provides a \textit{warm start} to the policy network but also improves upon the time to converge to good model architectures for given dataset (up to 5x reduction in time).
        Hence allowing us to get high performing compressed model architectures which satisfy user specific device thresholds, owing to the efficacy of transferring learnt information across dataset.
        Furthermore, we can train a policy for a larger dataset and subsequently fine tune the policy in a small data environment (often the case with data on mobile devices) to produce good architectures for the user tailored to the data.

\bibsubfile{named}{bibs/sub}

\section{Experimental Results}
    \label{sec:expts}
    We explain the relevant details of the experimental setup below.
    In the spirit of reproducibility, we have made all datasets, implementations for experiments, and results available\footnote{\url{http://bit.ly/2BTduY5}}.
    
    
    \subsection{Datasets}
        \label{sec:datasets}
        We use the CIFAR-10 and CIFAR-100~\cite{krizhevsky2009learning} datasets for our experiments.
        Both these datasets have 50,000 training images and 10,000 test images of size 32x32.
        They diff in the number of labelled classes, 10 vs 100.
        We will also work with specific subsets of these datasets (consisting of smaller number of class labels) for some of the experiments.
        We consider the following subsets:
        \begin{enumerate}
            \item   \texttt{Animals}: Subset of CIFAR-10 formed by the class labels \texttt{bird}, \texttt{cat}, \texttt{deer}, \texttt{dog}, \texttt{frog} and \texttt{horse}.
            \item   \texttt{Vehicles\textsubscript{10}}: Subset of CIFAR-10 formed by the class labels \texttt{airplane}, \texttt{automobile}, \texttt{ship} and \texttt{truck}.
            \item   For CIFAR-100, we use the superclasses as mentioned in the dataset.
            We use the superclasses \texttt{Insects}, \texttt{Fruits}, \texttt{Trees}, \texttt{Vehicles-1}, \texttt{Vehicles-2}, \texttt{People}, \texttt{Reptiles}.
        \end{enumerate}


    \subsection{Methods Under Study}
        \label{sec:baselines}
        \begin{enumerate}
            \item   \texttt{Prun}: We compare our method to the popular ranking based pruning method~\cite{molchanov2016pruning}.
            We remove 512 filters on each iteration of pruning heuristically followed by 10 epochs of fine-tuning to recover the network.
            \item   \texttt{KD}: Another popular approach for model compression is Knowledge Distillation.
            It requires a specific student architecture to train.
            We use a 7 layer deep CNN architecture inspired by the VGG architecture~\cite{simonyan2014very} for the student architecture.
            \item   \texttt{N2N}: This method systematically finds a compressed optimal architecture by searching within the teacher's architecture~\cite{ashok2017n2n}.
            The number of reinforcement learning iterations have been fixed at 100.
            In each iteration, 5 new student architectures are being trained using Knowledge distillation.
            \item   \texttt{AK}: Auto-Keras \cite{jin2018efficient} is a popular tool for architecture search which uses Bayesian optimization and searches a model from scratch given a dataset.
            \item   \texttt{DDC}: This is our proposed method.
        \end{enumerate}
    
    \subsection{Implementation details}
        \label{sec:ID}
        Our experiments provide evidence towards efficacy of imposing thresholds on reward to improve the architecture search.
        One important point to note here is that we do not use any proxies for inference time (like FLOPS) but instead we use actual inference time of the model on the system.
        For the memory footprint, we use number of parameters present in the model.
        Furthermore for consistency, we use the same teacher models across methods to provide evidence towards efficacy of our system.
        We use stochastic gradient descent with $momentum = 0.9 $ and $learning rate = 0.001$ for all our experiments.
        Unless otherwise mentioned the number of reinforcement learning iterations is 100 and in each iteration we sample 5 student architecture according to the RL policy and train each of these student architectures for 20 epochs.
    
    \subsection{Compressed Model Search}
        \label{sec:full-model-search}
        
    

\begin{table}[t]
	\centering
	\caption{Compressed Model Search for VGG11 on Cifar10}
	\begin{tabular}{lrrr} 
		\toprule
		Method~						& Accuracy~		& Compression~  & Inference Time (Sec)~	\\
		\midrule
		Original Model~				& 0.9~			& 1x~			& 3.43~			\\
		Pruning~					& 0.84~			& 5x~			& 1.96~			\\
		Knowledge Distillation~		& 0.79~			& 2x~			& 1.13~			\\
		Auto-Keras                  & 0.93			& 0.33x			& 7.9	\\
		N2N~						& \textbf{0.85}~& 9x~			& 1.49~			\\
		DDC$_{ht}$~					& 0.81~			& \textbf{20.8x}& 1.13~		\\
		DDC~	& 0.84~			& \textbf{20.8x}~& \textbf{1.11}~\\
		\bottomrule
	\end{tabular}
	\label{tab:VGG11-CIFAR10}
\end{table}

    \begin{table}[t]
    	\centering
    	\caption{Compressed Model Search for ResNet18 on Cifar10}
    	\begin{tabular}{lrrr}
    		\toprule
    		Method					& Acc.		    & Comp.	            & Inf.~Time (sec)	\\
    		\midrule
    		Original Model			& 0.83			& 1x				& 5.46			\\
    		Knowledge Distillation	& 0.80			& 1.67x				& 	3.12		\\
    		Auto-Keras              & 0.93			& 0.40x				& 7.9	\\
    		N2N						& \textbf{0.81}	& 4.34x				& 1.60			\\
    		DDC$_{ht}$				& \textbf{0.81}	& 2.56x				& 1.69			\\
    		DDC						& \textbf{0.81}	& \textbf{5.71}x	& \textbf{1.48}	\\
    		\bottomrule
    	\end{tabular}
    	\label{tab:ResNet18-CIFAR10}
    \end{table}

    \begin{table}[t]
    	\centering
    	\caption{Compressed Model Search for ResNet18 on Cifar100}
    	\begin{tabular}{lrrr}
    		\toprule
    		Method			& Acc.		    & Comp.		    & Inf.~Time (sec)	\\
    		\midrule
    		Original Model	& 0.72		    & 1x			& 2.71		\\
    		Auto-Keras      & 0.68			& 0.40x			& 6.2	\\
    		N2N				& 0.58		    & 4.1x			& 1.56		\\
    		DDC$_{ht}$      & \textbf{0.63} & 2.17x			& 1.68		\\
    		DDC				& 0.56			& \textbf{4.4x}	& \textbf{1.40}	\\
    		\bottomrule
    	\end{tabular}
    	\label{tab:ResNet18-CIFAR100}
    \end{table}

        In this set of experiments, we compress VGG11 and RESNET18 models for a good AMS trade-off on the entire dataset.
        We try all the methods detailed under Section~\ref{sec:baselines}.
        We consider 3 different combinations of architecture and dataset, \viz~VGG11 with CIFAR10, RESNET18 with CIFAR10, and RESNET18 with CIFAR100.
        In order to study the effect of soft targets on the AMS trade-off, we also learned the compression policy ($DDC_{ht}$) by training the student architecture with hard targets only during knowledge distillation.
        The results are in \tablesname~\ref{tab:VGG11-CIFAR10}, \ref{tab:ResNet18-CIFAR10} and~\ref{tab:ResNet18-CIFAR100}.
        
        The results indicate that our method is able to find highly compressed models that have low inference time without compromising much on the accuracy.
        Introduction of inference time as a metric in reward function not only helps the compression policy to find faster models but also helps the policy to achieve a much better accuracy-compression trade-off.
        In case of VGG11 with CIFAR10,  our method is able to find a compressed model that is 20.8x times smaller than the teacher model with a drop of 6\% accuracy.
        Also the produced compressed model is 3 times faster that the original model.
        We see that the accuracy tradeoff of compressed models in case of CIFAR100 is more when compared to models compressed on CIFAR10 dataset.
        This may be because CIFAR100 is a much harder dataset than CIFAR10.
        It has 100 classes with fewer samples per class than the CIFAR10 dataset. we also observe that the models produced by auto-keras though have good accuracy, are much larger that the original model thereby making them unusable for the task of model compression.
    \subsection{Subset based Compression}
        \label{sec:subset}
        
    \begin{table}[t]
        \centering
        \caption{ResNet18 AMS Trade-off of DDC on full vs partial CIFAR100 dataset}
        \begin{tabular}{llrrr}
            \toprule
            Subset                      & Policy    & Acc.  & Comp. & Inf.~Time (sec)   \\
            \midrule
            \multirow{2}*{fruits}       & Full      & 0.64  & 4.34x & 0.12  \\
                                        & Sub       & \textbf{0.73}  & \textbf{6.3x} & \textbf{0.096} \\
            \midrule
            \multirow{2}*{insects}      & Full      & 0.61  & \textbf{4.3x} & 0.12  \\
                                        & Sub       & \textbf{0.67}  & 3.3x & \textbf{0.087} \\
            \midrule
            \multirow{2}*{people}       & Full      & 0.36  & 4.3x & 0.11  \\
                                        & Sub       & \textbf{0.388} & \textbf{6.3x} & \textbf{0.064} \\
            \midrule
            \multirow{2}*{reptiles}     & Full      & 0.32  & 4.3x & 0.11  \\
                                        & Sub       & \textbf{0.556} & \textbf{4.7x} & \textbf{0.069} \\
            \midrule
            \multirow{2}*{trees}        & Full      & 0.57  & \textbf{4.3x} & 0.11  \\
                                        & Sub       & \textbf{0.606} & 3.4x & \textbf{0.099} \\
            \midrule
            \multirow{2}*{vehicles-1}   & Full      & \textbf{0.67}  & \textbf{4.3x} & 0.13  \\
                                        & Sub       & 0.664 & 2.3x & \textbf{0.079} \\
            \midrule
            \multirow{2}*{vehicles-2}   & Full      & 0.57  & 4.3x & 0.13  \\
                                        & Sub       & \textbf{0.765} & \textbf{6.3x}  & \textbf{0.064} \\
            \bottomrule
        \end{tabular}
        \label{tab:resnet18-cifar100-fulltosub}
    \end{table}

    \begin{table}[t]
        \centering
        \caption{ResNet18 AMS Trade-off of DDC on full vs partial CIFAR10 dataset}
        \begin{tabular}{llrrr}
            \toprule
            Subset                          & Policy    & Acc.          & Comp.         & Inf.~Time (sec)   \\
            \midrule
            \multirow{2}*{animals}          & Full      & 0.79          & \textbf{5.7x} & 0.99              \\
                                            & Sub       & \textbf{0.815}& 4.14x         & \textbf{0.913}    \\
            \midrule
            \multirow{2}*{vehicles$_{10}$}  & Full      & 0.85          & 5.71x         & 0.703             \\
                                            & Sub       & \textbf{0.89} & \textbf{5.89x}& \textbf{0.549}    \\
            \bottomrule
        \end{tabular}
        \label{tab:resnet18-cifar10-fulltosub}
    \end{table}

    \begin{table}[t]
        \centering
        \caption{VGG11 AMS Trade-off of DDC on full vs partial CIFAR10 dataset}
        \begin{tabular}{llrrr}
            \toprule
            Subset                          & Policy    & Acc.          & Comp.         & Inf.~Time (sec)   \\
            \midrule
            \multirow{2}*{animals}          & Full      & \textbf{0.79} & \textbf{20x}  & 0.862             \\
                                            & Sub       & \textbf{0.79} & 14.3x         & \textbf{0.75}     \\
            \midrule
            \multirow{2}*{vehicles$_{10}$}  & Full      & \textbf{0.9}  & 20x           & 0.623             \\
                                            & Sub       & 0.87          & \textbf{50x}  & \textbf{0.415}    \\
            \bottomrule
        \end{tabular}
        \label{tab:vgg11-cifar10-fulltosub}
    \end{table}

        If high accuracy is needed only on a subset of the class labels, it is possible to imagine that much better inference speed and memory footprints may be possible post-compression.
        In subset based compression we learn the RL policy by training the student architectures sampled in each RL iteration, only on the desired subsets rather than on the entire dataset.
        This will enable our compression method to learn subset specific policy that has a better AMS tradeoff when compared to the policy learnt on the entire dataset.
        
        Tables~\ref{tab:resnet18-cifar100-fulltosub}, \ref{tab:resnet18-cifar10-fulltosub} and \ref{tab:vgg11-cifar10-fulltosub} compares the performance of the compressed models produced by policies learnt on subsets with the compressed model produced by the policy learnt on the entire dataset.
        The results indicate that the models produced by the polices learnt on the subsets outperforms the model produced by the policy learned on the entire dataset on all 3 performance metrics for most of the subsets.
        The enhancement in performance is clearly evident in the case of CIFAR-100 subsets.
        The policy learns to remove layers of the teacher model who's learned features doesn't help in differentiating samples from the subsets thus, giving a better AMS tradeoff.
    
    \subsection{Transfer of Compression Policy}
    \label{sec:poltrans}
        The previous experiment demonstrates that subset based model compression helps in achieving a better AMS tradeoff.
        But datasets for practical applications are often huge with large number of classes.
        Finding compressed policy for all the subsets of classes will be prohibitive.
        In terms of concrete numbers, learning a policy for the animals subset of CIFAR-10 using RESNET18 as teacher model takes around 12 hours for DDC.
        We propose compression policy transfer as a way to scale our algorithm for large scale deployments.
        In policy transfer, we use the compression policy learnt on the entire dataset to bootstrap the learning of the policy for the subsets.
        We observed that the compression policy learnt by policy transfer is able to produce compressed models with a good AMS tradeoff in around 20 RL iterations.
        
    \begin{table}[t]
        \centering
        \caption{ResNet18 AMS Trade-off of DDC on policy transfer vs learning from scratch on subsets of CIFAR100 dataset after 20 epochs}
        \begin{tabular}{llrrr}
            \toprule
            Subset                      & Policy    & Acc.  & Comp. & Inf.~Time (sec)   \\
            \midrule
            \multirow{2}*{fruits}       & Transfer  & \textbf{0.748} & \textbf{6.3x} & \textbf{0.062}  \\
                                        & Scratch   & 0.716 & 4.3x & 0.109  \\
            \midrule
            \multirow{2}*{insects}      & Transfer  & 0.646 & \textbf{6.3x} & \textbf{0.063}  \\
                                        & Scratch   & \textbf{0.66}  & 2x & 0.079  \\
            \midrule
            \multirow{2}*{people}       & Transfer  & \textbf{0.393} & \textbf{6.3x} & \textbf{0.066}  \\
                                        & Scratch   & 0.354 & 4.7x & 0.073  \\
            \midrule
            \multirow{2}*{reptiles}     & Transfer  & \textbf{0.566} & \textbf{6.3x} & \textbf{0.065}  \\
                                        & Scratch   & 0.534 & 2x & 0.081  \\
            \midrule
            \multirow{2}*{trees}        & Transfer  & \textbf{0.592} & \textbf{4.7x} & \textbf{0.067}  \\
                                        & Scratch   & 0.568 & 3.4x & 0.085  \\
            \midrule
            \multirow{2}*{vehicles-1}   & Transfer  & \textbf{0.678} & \textbf{4.7x} & \textbf{0.069}  \\
                                        & Scratch   & 0.62  & 1.7x & 0.089  \\
            \midrule
            \multirow{2}*{vehicles-2}   & Transfer  & \textbf{0.79}  & \textbf{6.3x} & \textbf{0.063}  \\
                                        & Scratch   & 0.714 & 4.3x & 0.083  \\
            \bottomrule
        \end{tabular}
        \label{tab:resnet18-cifar100-pt}
    \end{table}

    \begin{table}[t]
        \centering
        \caption{ResNet18 AMS Trade-off of DDC on policy transfer vs learning from scratch on subsets of CIFAR10 dataset after 20 epochs}
        \begin{tabular}{llrrr}
            \toprule
            Subset                          & Policy    & Acc.          & Comp.         & Inf.~Time (sec)   \\
            \midrule
            \multirow{2}*{animals}          & Transfer  & 0.783         & \textbf{6.45x}& \textbf{0.801}    \\ 
                                            & Scratch   & \textbf{0.829}& 3.5x          & 0.968             \\ 
            \midrule
            \multirow{2}*{vehicles$_{10}$}  & Transfer  & 0.888         & \textbf{6.45x}& \textbf{0.496}    \\
                                            & Scratch   & \textbf{0.896}& 5.8x          & 0.580             \\
            \bottomrule
        \end{tabular}
        \label{tab:resnet18-cifar10-pt}
    \end{table}

    \begin{table}[t]
        \centering
        \caption{VGG11 AMS Trade-off of DDC on policy transfer vs learning from scratch on subsets of CIFAR10 dataset after 20 epochs}
        \begin{tabular}{llrrr}
            \toprule
            Subset                          & Policy    & Acc.          & Comp.             & Inf.~Time (sec)   \\
            \midrule
            \multirow{2}*{animals}          & Transfer  & \textbf{0.81} & \textbf{20.83x}   & \textbf{0.69}     \\ 
                                            & Scratch   & 0.81          & 7.57x             & 0.732             \\ 
            \midrule
            \multirow{2}*{vehicles$_{10}$}  & Transfer  & 0.880         & \textbf{43.4x}    & \textbf{0.423}    \\
                                            & Scratch   & \textbf{0.894}& 7.81x             & 0.425             \\
            \bottomrule
        \end{tabular}
        \label{tab:vgg11-cifar10-pt}
    \end{table}

        Tables~\ref{tab:resnet18-cifar100-pt}, \ref{tab:resnet18-cifar10-pt} and \ref{tab:vgg11-cifar10-pt} tabulates the performance of the compressed models produced by policies learnt from scratch and policies learnt from policy transfer after 20 RL iterations.
        The performance of the policy transfered models after 20 epoch are comparable to policy learnt from scratch models after 100 epoch.

    \bibsubfile{named}{bibs/sub}

\section{Conclusion}
    \label{sec:conclude}
    This paper presents an automated systematic approach to compress a convolutional neural networks using Reinforcement learning.
    The key idea behind this method is to incorporate real-world latency information into reward function to find faster performing compressed models without much tradeoff in accuracy. We demonstrated that subset based model compression can be helpful in scenarios where the deployed model sees only a subset of data classes. We proposed policy transfer as way to solve large scale deployment of our compression algorithm.


\bibliographystyle{ACM-Reference-Format}
\bibliography{IEEEabrv,bibs/sub}

\end{document}